\documentclass[conference]{IEEEtran}
\IEEEoverridecommandlockouts
\usepackage{amsmath,amssymb,amsfonts}
\usepackage{algorithmic}
\usepackage{graphicx}
\usepackage{textcomp}
\usepackage{xcolor}
\usepackage{hyperref}
\usepackage{biblatex}
\usepackage{soul}
\usepackage{balance}
\usepackage[normalem]{ulem} 
\usepackage{makecell}
\usepackage{combelow}
\usepackage[normalem]{ulem}
\usepackage{multirow}

\def\BibTeX{{\rm B\kern-.05em{\sc i\kern-.025em b}\kern-.08em
    T\kern-.1667em\lower.7ex\hbox{E}\kern-.125emX}}
\begin{document}

\title{An Analysis of Multi-Task Architectures for the Hierarchic Multi-Label Problem of Vehicle Model and Make Classification}

\author{\IEEEauthorblockN{Alexandru Manole, Laura-Silvia Dio\cb{s}an}}

\author{
    \IEEEauthorblockN{Alexandru Manole\IEEEauthorrefmark{1} and Laura-Silvia Dio\c{s}an\IEEEauthorrefmark{1}}
    \IEEEauthorblockA{\IEEEauthorrefmark{1}Department of Computer Science, Babe\c{s}-Bolyai University, Cluj-Napoca, Romania \\
    \{alexandru.manole, laura.diosan\}@ubbcluj.ro}
}

\maketitle

\begin{abstract}
Most information in our world is organized hierarchically; however, many Deep Learning approaches do not leverage this semantically rich structure. Research suggests that human learning benefits from exploiting the hierarchical structure of information, and intelligent models could similarly take advantage of this through multi-task learning. In this work, we analyze the advantages and limitations of multi-task learning in a hierarchical multi-label classification problem: car make and model classification. Considering both parallel and cascaded multi-task architectures, we evaluate their impact on different Deep Learning classifiers (CNNs, Transformers) while varying key factors such as dropout rate and loss weighting to gain deeper insight into the effectiveness of this approach. The tests are conducted on two established benchmarks: StanfordCars and CompCars. We observe the effectiveness of the multi-task paradigm on both datasets, improving the performance of the investigated CNN in almost all scenarios. Furthermore, the approach yields significant improvements on the CompCars dataset for both types of models.

\end{abstract}


\section{Introduction}
\label{sec:intro}

\IEEEPARstart{I}{n} the past decade Deep Learning approaches have been used successfully in numerous fields for a myriad of tasks. Usually, the training process focuses on a single task, in which the intelligent model becomes a specialist. Although this method can yield impressive performance, some limitations still restrict the performance of this technique. One such deficiency is related to the generalization capability of Deep Learning solutions, as the model can yield surprising and wrong predictions for out-of-distribution samples. 

In our world, many entities can be observed as part of a hierarchical structure. For example, in Biology all discovered species are organized using taxonomies with around 10 hierarchical levels, ranging from: \textit{Domain}, \textit{Kingdom} to \textit{Genus} and \textit{Species}. The stratified structure allows scientists to increase their understanding of each species. In addition to acting as a logical organizational form for specialists, hierarchical structures are considered an essential construction in the learning process, as shown by cognitive scientists \cite{botvinick2009hierarchically}, \cite{eckstein2021mind}, \cite{theves2021learning}. 



Our hypothesis, which is also supported by innovative works in the literature \cite{chen2019deep}, \cite{pujari2021multi}, \cite{wang2021label},  \cite{wang2023consistency}, \cite{jiang2024hierarchical}, argues that deep learning model training benefits from the incorporation of additional information regarding the super-classes or sub-classes of the target classification task. These approaches apply a variety of techniques, in order to fully exploit the stratified label organization, including Multi-task Learning (\textbf{MTL}), a paradigm in which multiple objectives are combined in a single training procedure. MTL is a suitable choice for Hierarchic Multi-Label problems, as for each level from the taxonomy, a classification head can be created. Depending on the relation between the coarse classification and the fine-grained classification, we can have parallel objectives which do not interact directly or cascaded objectives. In the latter, the result of the first prediction is used as an additional input for the following tasks.

Although multitask models are applied for hierarchical classification problems, the effects of the MTL architecture choice have yet to be thoroughly explored. Starting from the Vehicle Make and Model Classification (\textbf{VMMC}) problem, a challenge in which the intelligent system has to jointly predict the car make (i.e. 'Dacia') and model (i.e. 'Logan'), this paper investigates the behavior of MTL architectures. 
The novel aspects of this work for the VMMC are: the investigation of the transformer architecture, the exploration of the influence of loss weights in the MTL context and an in-depth analysis of MTL structures, using multiple base networks and datasets. This work aims to provide answers to the
following research questions.



\begin{enumerate}
    \item[\textbf{RQ1}:] 
    How might the addition of cascaded and parallel MTL objectives influence the VMMC problem?
    \item[\textbf{RQ2}:] 
    How does the architecture of the base model (CNN, Transformer) influence the effectiveness of the multi-task model?
\end{enumerate}


This work brings several novel contributions to the Vehicle Make and Model Classification task. 

First, we conduct an extensive experimental analysis in both single-task (ST) and multi-task (MT) learning setups, exploring a variety of architectures and combined loss functions. To our knowledge, such a comprehensive comparative analysis, especially involving MT with different task weights and dropout values, has not been previously reported. Second, we investigate a broad range of convolutional neural networks (CNNs), extending prior work such as \cite{avianto2022cnn} and \cite{lyu2022framework} by experimenting with a more diverse set of CNN variants. Furthermore, we explore the use of Vision Transformers (ViTs), which, to the best of our knowledge, have not yet been thoroughly investigated in the VMMC task. Our experiments include multiple recent and state-of-the-art ViT architectures, which introduce a novel direction in this domain. Lastly, we evaluate our approaches on two large and diverse real-world datasets, Stanford Cars and CompCars, unlike many existing studies that rely on a single dataset. While some previous works \cite{liu2024progressive}, \cite{yu2022embedding} evaluated on multiple datasets, the former aimed at broader multitask learning applications, and the latter focuses solely on a single-task approach, making our evaluation setup more robust and focused on the specific challenges of VMMC.

The structure of the rest of this paper is the following: \textbf{Section} \ref{relWork} describes the related methods used to solve this task or other similar problems, \textbf{Section} \ref{sec:methodology} presents the proposed approach for the vehicle model and make classification task and \textbf{Section} \ref{results} describes the proposed experiments and the analysis of their results. The paper is concluded in \textbf{Section} \ref{conclusion}.

\section{Backgrounds and Related Work}
\label{relWork}

\subsection{Multi-Task Learning}

Multi-Task is a Machine Learning paradigm introduced in \cite{caruana1993multitask} \cite{caruana1997multitask} that aims to increase the performance and generalization capabilities of intelligent models through the addition of symbiotic objectives. This paradigm  
design philosophy proved itself useful in the deep learning era, achieving impressive results in both Computer Vision and Natural Language Processing \cite{zhang2021survey}. 

The goal of multi-task learning is to solve related but distinct tasks at the same time by using task coupling, which enhances the learning process overall and produces better final models. Multitask learning has been defined in a variety of ways \cite{caruana1997multitask}, but generally speaking, its primary objective is to improve generalization performance by using an inductive transfer mechanism that makes use of the domain-specific knowledge found in the training data of related problems. This is accomplished by employing a common representation to learn various tasks concurrently. Following the formulation of \cite{baxter2000model}, to formalize MTL we have to consider: 

\begin{itemize}
    \item a finite set of $m$ tasks
    \item an input space $X$ (e.g. a set of images)
    \item an output space $(Y_1, Y_2, \ldots, Y_m)$ (e.g. $m$ sets of labels)
    \item $m$ probability distributions $P = (P_1, P_2, \ldots, P_m)$, $P_t$ on $X \times Y_t$, $t \in \{1, 2, \ldots, m\}$
    \item $m$ loss functions $(loss_1, loss_2, \ldots, loss_m)$, $loss_t : Y_t \times Y_t \rightarrow \Re$, $t \in \{1, 2, \ldots, m\}$
    \item a hypothesis space $(H_1, H_2, \ldots, H_m)$ which is a set of functions $h_t : X \rightarrow Y_t$, $t \in \{1, 2, \ldots, m\}$
\end{itemize}

The goal of the MTL algorithm is to select a set of $m$ hypotheses $h_t \in H$ with minimal expected loss: 
\begin{equation}
    err_P(h) = \sum_{t=1}^{m} \int_{X \times Y_t} loss (h_t(x), y_t) \,dP(x, y_t)
    \label{eq:expectedLoss}
\end{equation}
    
Given a training dataset: 
\begin{equation} 
    \begin{split}
    z = \{  & (x^1, y^1_1, y^1_2, \ldots y^1_m), \\
            & (x^2, y^2_1, y^2_2, \ldots y^2_m), \\
            & \ldots, \\ 
            & (x^n, y^n_1, y^n_2, \ldots y^n_m)\}
    \end{split}
    \label{e:trainData}
\end{equation}

\noindent
many algorithms seek to minimize the empirical loss defined by: 
\begin{equation} 
    \begin{split}
    err_z(h) =  & \frac{1}{n}\sum_{i=1}^{n} loss(h_1(x^i),y_1^i)  + \\
                & \frac{1}{n} \sum_{i=1}^{n} loss(h_2(x^i),y_2^i) + \\
                & \ldots, \\ 
                & \frac{1}{n} \sum_{i=1}^{n} loss(h_m(x^i),y_m^i)
    \end{split}
    \label{eq:loss}
\end{equation}

The $m$ tasks could be homogeneous (e.g., various classification tasks) or heterogeneous (e.g. regression tasks mixed by classification tasks). 
We shall limit our work to deep learning-based methods and homogeneous tasks. 

\begin{figure}[hbt!]
\centering
\centerline{\includegraphics[scale=0.3]{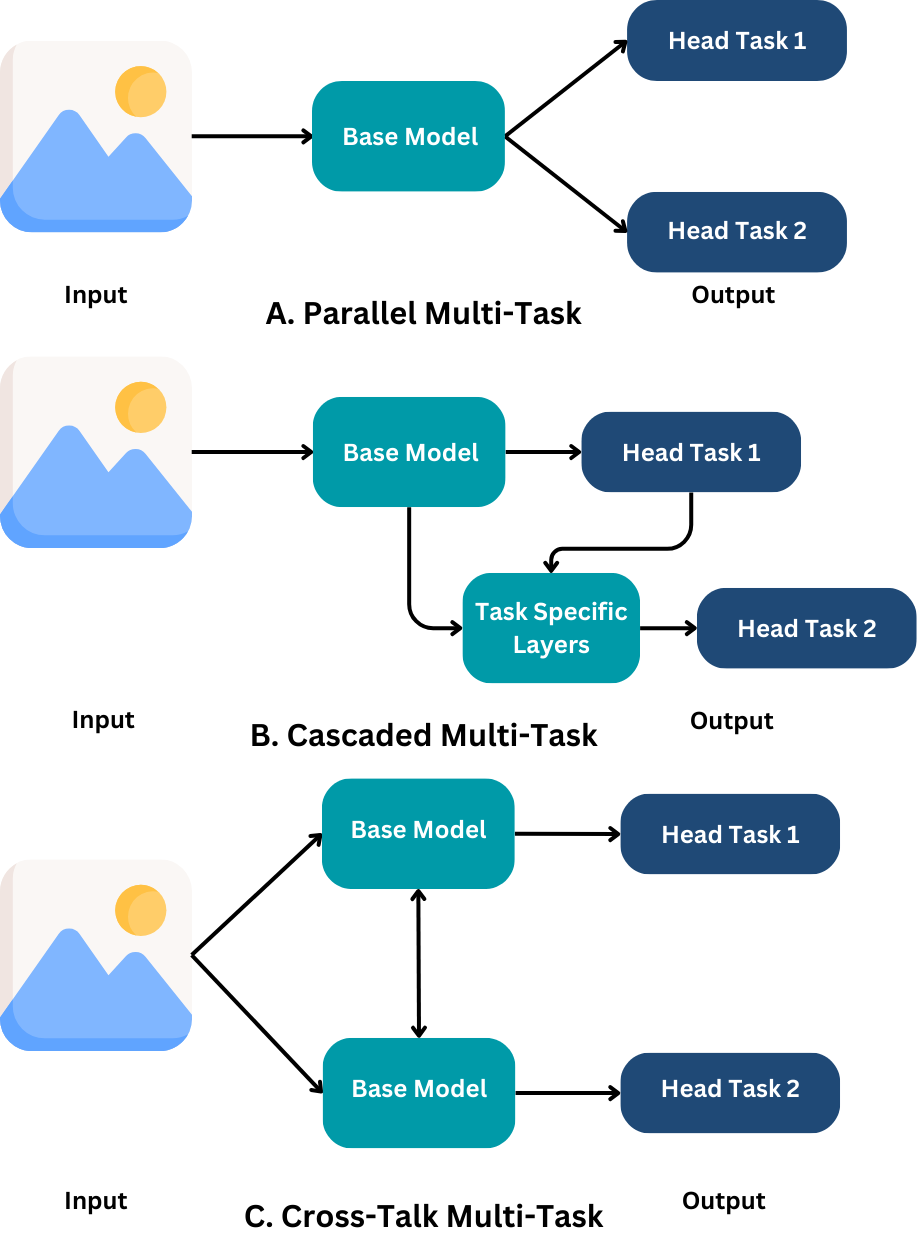}}
	\caption{Simplified overview of multi-task architectures. (A) Parallel (B) Cascaded (C) Cross-Talk Multitask.}
	\label{fig:multi-task-taxonomy}
\end{figure}

Depending on the way the task objectives and their features are combined, three main Multi-Task architectures exist: parallel, cascaded, and cross-talk 
\cite{crawshaw2020multi} \cite{zhang2021survey}.
In the former, all tasks share the same encoder, which feeds its features into task-specific layers for each classification objective. In the cascaded MTL, one of the tasks is initially solved, and some of the obtained features are used as part of the other tasks. The cross-talk architecture resembles the parallel one, the main difference being that in this structure two task-specific paths are allowed to communicate and share features through fusion modules. A simplified illustration of MTL architectures, composed of only two tasks is presented in \textbf{Figure} \ref{fig:multi-task-taxonomy}.

In its initial form, MTL was introduced as a parallel architecture by Caruana \cite{caruana1993multitask} \cite{caruana1997multitask}. Starting from a single input, a series of common layers are used to extract relevant features that are then fed into a final set of task-specific dense layers. Each task has only a single task-specific layer, thus most parameters are shared between al objectives. This is known today as hard-parameter sharing multitask, as opposed to alternative soft-parameter sharing architectures in which the number of task-specific parameters is higher.

Soft-parameter sharing parallel networks were improved by adding modules that allow the sharing of information between previously independent tasks, resulting in the creation of cross-talk (interacted) MTL. This concept was introduced in Natural Language Processing  in \cite{duong2015low}  and in Computer Vision by \cite{misra2016cross}.

The cascaded approach was established in Computer Vision in works such as \cite{dai2016instance} and \cite{zhang2016joint} due to its hierarchical process where tasks are solved sequentially. Using the output of a task as a feature for the following tasks results in a smoother learning curve, which is shown to improve both efficiency and accuracy. As shown in \cite{zhang2021survey}, MTL has certain similarities to other paradigms such as multi-view and multi-label learning, but the most notable one is between cascaded multi-task and transfer learning (\textbf{TL}) \cite{pan2010transfer}. Although MTL is often used to obtain two or more distinct outputs, the cascaded variant of the paradigm is also used to approach multiple tasks with the purpose of refining a principal task. TL also uses a source task, usually with more data available, to increase the performance of a target task, but in this case two distinct datasets are used. 

In the case of homogeneous tasks as classification,  cross-entropy loss can be used for each task and weight them with hyper-parameters 
Other strategies were attempted to overcome the need for this computationally expensive search, but alternatives like: Uncertainty-Based Loss Weighting \cite{kendall2018multi} and Dynamic Task Prioritization \cite{guo2018dynamic} did not yield fruitful results in this case.

\begin{table}[htbp]
    \centering
    \begin{tabular}{|c|c|c|c|c|c|c|c|}
        \hline
        \multirow{2}{*}{Ref} & \multicolumn{3}{c|}{Tasks} & \multirow{2}{*}{MTL} & \multicolumn{3}{c|}{Datasets}\\ \cline{2-4} \cline{6-8}
         & make & model & other &  & \makecell{\textbf{Stanford}\\\textbf{Cars}} & \makecell{\textbf{Comp}\\\textbf{Car}} & Other\\\hline
        \cite{kemertas2020rankmi} & & \checkmark & & & \checkmark & &\\ \hline
        \cite{yu2022embedding} & & \checkmark & & & \checkmark & \checkmark &\\ \hline
        \cite{liu2024progressive} & & \checkmark & \makecell{image\\denoising} & \checkmark & \checkmark & \checkmark & \checkmark \\ \hline
        \cite{dai2017efficient} & & \checkmark & \makecell{car\\localization} & \checkmark & \checkmark & & \\ \hline        
        \cite{xia2016vehicle} & \checkmark & & \makecell{attributes\\prediction} & \checkmark & & & \checkmark \\ \hline
        \cite{sun2019vehicle} & \checkmark & & \makecell{car\\colour} & \checkmark & & & \checkmark \\ \hline
        \cite{avianto2022cnn} & \checkmark & \checkmark & & \checkmark & & & \checkmark \\ \hline
        \cite{lyu2022framework} & \checkmark & \checkmark & & \checkmark & & & \checkmark \\ \hline
        \textbf{ours} & \checkmark & \checkmark & & \checkmark & \checkmark & \checkmark & \\ \hline
    \end{tabular}
    \caption{Comparison of related work for make and model vehicle recognition}
    \label{tab:MMVRcomparisonMTL}
\end{table}



\begin{table*}[ht]
    \centering
    \begin{tabular}{|c|c|c|c|c|c|c|c|}
        \hline
        \multirow{3}{*}{Ref} & \multicolumn{3}{c|}{Architecture} & \multicolumn{4}{c|}{Learning} \\ \cline{2-8}
        & \multirow{2}{*}{Base model} & \multicolumn{2}{c|}{Heads} & \multirow{2}{*}{Pre-trained} & \multirow{2}{*}{Loss} & \multirow{2}{*}{Fusion} & \multirow{2}{*}{Process} \\ \cline{3-4}
        & & Make & Model & & & & \\ \hline
        
        \cite{avianto2022cnn} & VGG & FCL & FCL & No & CE & \makecell{Weighted \\ (model vs. make)} & End-to-end \\ \hline
        
        \cite{lyu2022framework} & \makecell{AlexNet\\ResNet50\\DenseNet121} & -- & FCL & Yes & CE & \makecell{Unweighted \\ (model only)} & Not end-to-end \\ \hline
        Proposed & 
        
        \makecell{MobileNet\\ResNet\\DenseNet\\EfficientNet\\ViT-B\\Swin\\MaxViT\\ConvNeXt} & FCL & FCL & Yes & CE & \makecell{Weighted \\ (model vs. make)} & End-to-end \\ \hline
        
    \end{tabular}
    \caption{Comparison of related work for MTL for make and model vehicle recognition – architecture and learning strategies}
    \label{tab:MMVRcomparison}
\end{table*}

\subsection{Vehicle Make and Model Classification}
\label{sec:VMMR}

The image classification task known as \textit{hierarchical image classification} involves providing the image together with hierarchical information about the classifications. The hierarchy information is typically described as a tree structure with classes as nodes: the leaf nodes of the tree have a high level of detail and are called fine classes, while the other nodes are called coarse classes \cite{bertinetto2020making}. 
The hierarchy information can be integrated in the learning algorithms and used to increase the accuracy of image recognition. However, agnostic MTL methods to the hierarchy of labels are characterized by simplicity and flexibility (they can be applied to various datasets). They often generalize well across different tasks since they treat each label independently. Furthermore, they do not face the scalability problem of the methods integrating hierarchy information in the learning process.

A hierarchical classification problem that has been extensively studied \cite{gayen2024two} is the classification of vehicle make and model. In addition to its numerous applications in intelligent transport systems, the challenges that come with this task make it a suitable benchmark to investigate deep learning approaches. 

\begin{figure*}[hbt!]
\centering
\centerline{\includegraphics[scale=0.65]{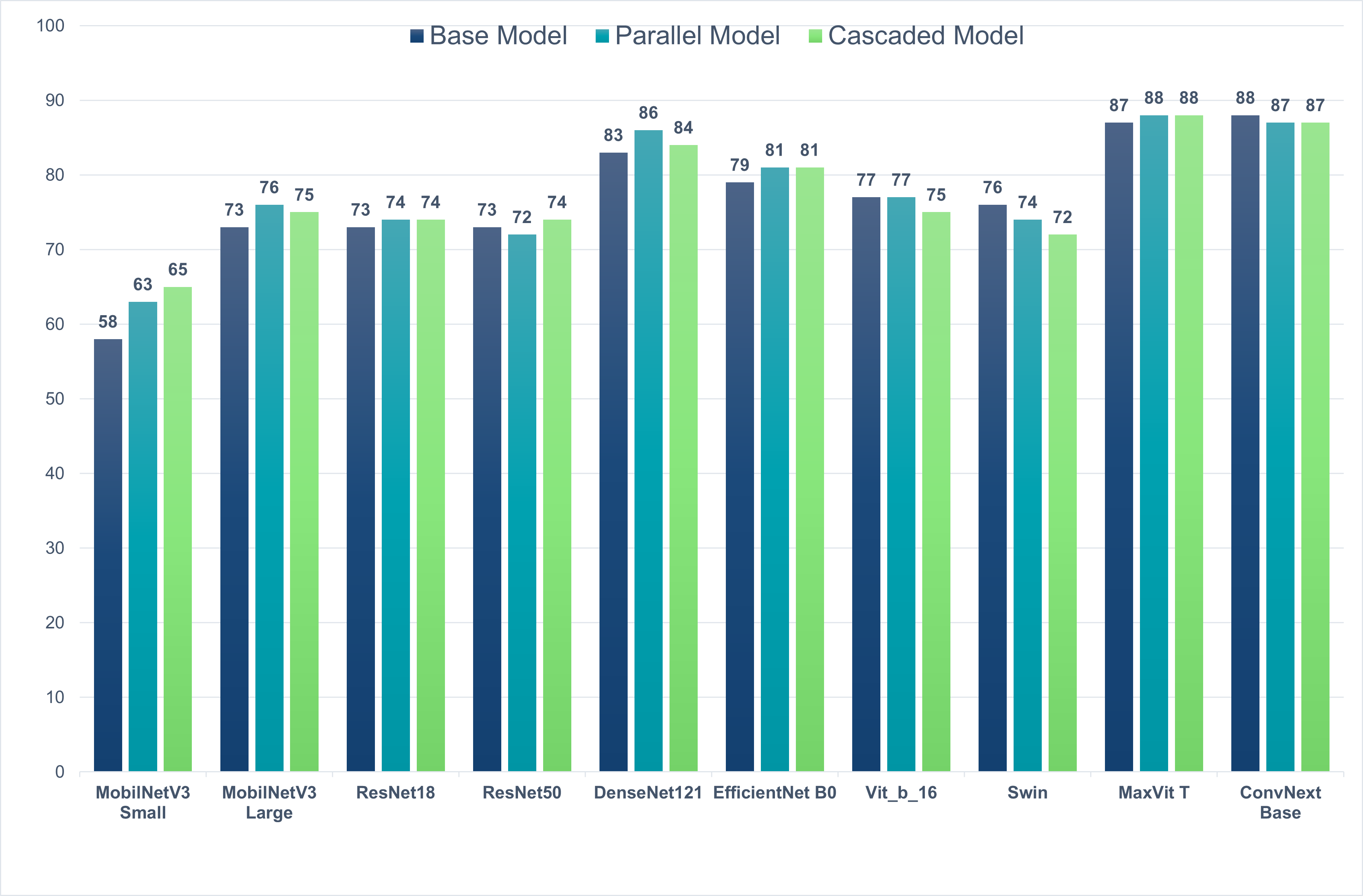}}
	\caption{Test accuracy for car model prediction for the initial experiment used for choosing the most promising base models on Stanford Cars (no dropout, $\lambda_1 = 0.9$, $\lambda_2=0.1$).}
	\label{fig:experiment0}
\end{figure*}

Different car models produced by the same manufacturer can present similar characteristics, leading to high inter-class similarity. This issue is further aggravated by the fact that all car manufacturers produce the same car topologies: SUVs, Sedans, Hatchbacks, etc. Thus, even models from different makers can resemble each other. Furthermore, intra-class variance is another issue which limits the performance of deep-learning approaches on this task. First of all, the same car model can look drastically different in two different instances as some elements of the car can be customized (paint color, car rims). Secondly, occlusions, weather conditions, lighting variations, and other environmental factors can cause images of the same vehicle to appear different. 

The literature consists of an impressive amount of work which tackles the vehicle model and make classification. In \cite{gayen2024two} a taxonomy of these approaches is proposed and three main architectural types are identified: convolutional neural networks (\textbf{CNNs}), recurrent neural networks (\textbf{RNNs}) and Attention-based models. Currently, CNNs achieved the highest accuracy on the two main VMMC benchmarks: Stanford Cars \cite{krause2015fine} and CompCars \cite{yang2015large}. 


Until recently, two different approaches achieved the best performance on the datasets. Kamertas et al. \cite{kemertas2020rankmi} proposed a novel loss function, dubbed RankMI, which similalary to Triplet Loss attempts to increase the similarity of embeddings obtained from images of the same class. Through the use of Mutual Information Maximizing, the distributions of matching and non-matching pair are disentangled, leading to better image retrieval. Although it was not directly proposed for classification, this approach can be used to obtain competitive results on Stanford Cars, reaching an accuracy of around 96. 5\%. 

In \cite{yu2022embedding}, the authors proposed a novel way to embed pose and perspective information into the classifier. Their approach can be divided into two sub-networks: the pose estimation CNN (\textbf{PE-CNN}), a YoloV3 \cite{redmon2018yolov3} which jointly estimates the pose and classifies the view (rear, front, side) and a Vehicle Model Classification CNN (\textbf{(VMC-CNN}). The latter enhances a ResNet50 \cite{he2016deep} with features obtained by the PE-CNN sub-network. This rich representation yields an accuracy of 98.9\% on the CompCar.

Another multi-task approach, introduced in \cite{liu2024progressive}, was recently able to exceed the performance of the methods mentioned above, reaching accuracies of 97.3\% in the Stanford Cars dataset and 99.1\% in CompCars. They are able to obtain these impressive results through one novel frameworks: progressive multi-task antinoise learning (\textbf{PMAL}), which combines the recognition task with the self-supervised image de-noising task.

\begin{figure*}[hbt!]
\centering
\centerline{\includegraphics[scale=0.65]{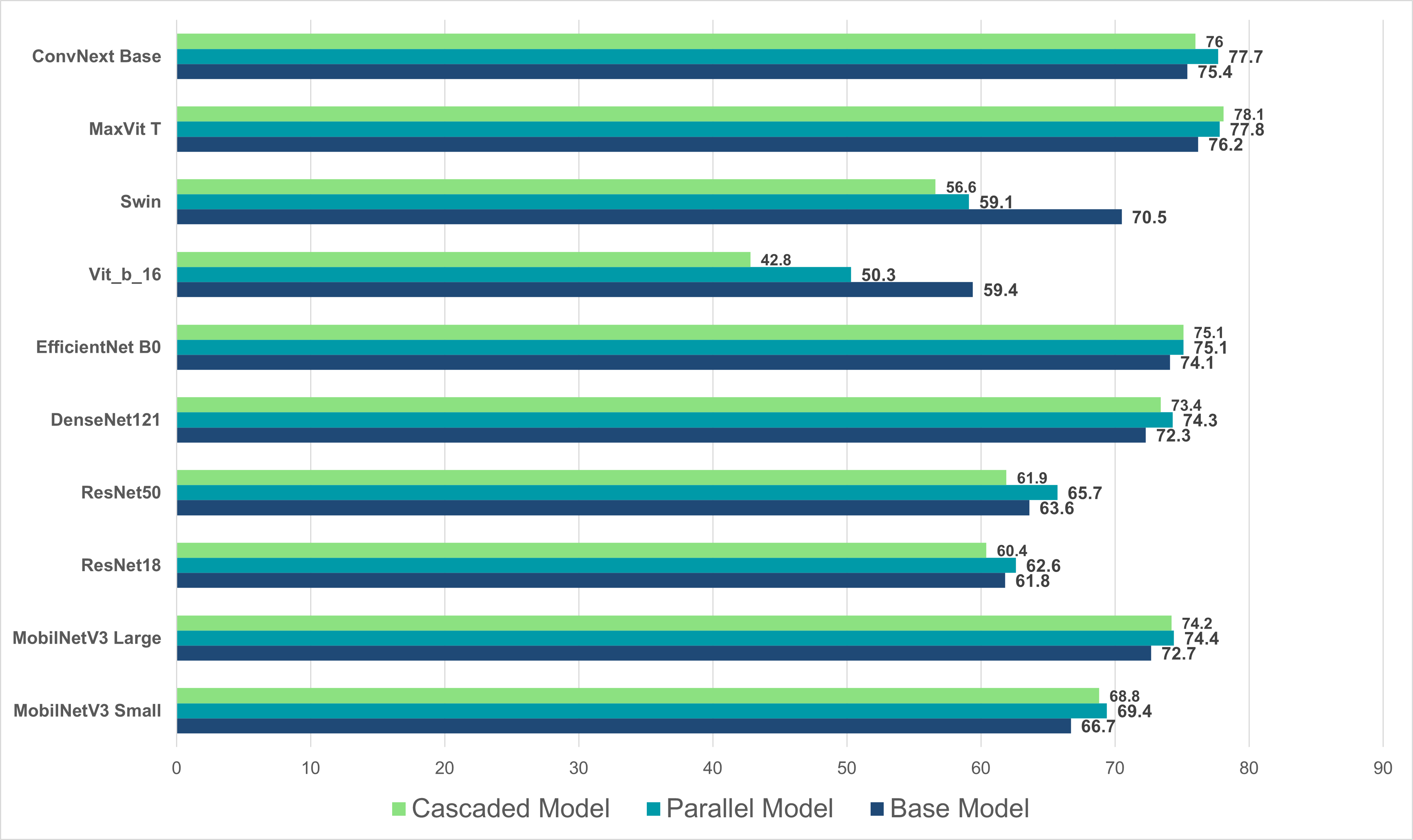}}
	\caption{Test accuracy for car model prediction for the initial experiment used for choosing the most promising base models on CompCars (no dropout, $\lambda_1 = 0.9$, $\lambda_2=0.1$).}
	\label{fig:experiment0_compcars}
\end{figure*}

Various other MTL frameworks have been proposed throughout the years. \cite{xia2016vehicle} and \cite{yu2019cascaded} attempt to recognize the vehicle's brand by correctly classifying the logo. The former paper introduces a hard-parameter sharing parallel architecture which simultaneously predicts the logo's class and 9 other attributes which describe it including: X-axis symmetry, Y-axis symmetry, presence of an animal mascot, and country of origin. 

In \cite{yu2019cascaded}, the logo localization is performed jointly with the classification task as a multi-task cascaded optimization problem in which a region-proposal networks feeds crops of the image into the classifier module. Dai et al. \cite{dai2017efficient} also combine localization and classification but with the purpose of correctly identifying the car model from images of the entire car, such as those from the Stanford Cars dataset. Starting from a VGG16, which is used as an image encoder, a rich feature map is obtained which is fed into two parallel sub-networks: the first one specialized in part localization, able to find key-points and another which predict the class label. 

Other works in the literature combine two classification tasks in an MTL framework. The approach from \cite{sun2019vehicle} can predict both the brand and the colour of the car through the use of a parallel architecture with soft-parameter sharing built on top of a DenseNet base model. The method is tested on the PKU-VD Dataset \cite{yan2017exploiting} and yields over 99\% in brand accuracy. Avianto et al. \cite{avianto2022cnn} predict the model and make simultaneously using a DenseNet extended with two parallel branches, each having 3 dense layers, with dropout layers between them. On the InaV-Dash they achieve 98.73\% brand accuracy and 97.69\% vehicle accuracy. The existing approaches are summarized in \textbf{Table} \ref{tab:MMVRcomparisonMTL}.


\begin{table}[ht]
    \centering
    \scriptsize
    \caption{Number of parameters for each base and multi-task model.}
    \scalebox{1.2}{ 
    \begin{tabular}{||c c c c||} 
     \hline
     \textbf{Model} & \textbf{MTL} & \textbf{No. Parameters} & \textbf{FLOPs}\\ 
     \hline\hline
     DenseNet-121 & No & 7,154,756 &  2,896,183,808 \\ 
     \hline
     DenseNet-121 & Parallel & 7,204,981 & 2,896,233,984  \\
     \hline
     DenseNet-121 & Cascaded & 7,214,585 & 2,896,243,588 \\
     \hline\hline
     MaxVit-T & No & 30,508,172 & 5,455,904,000 \\
     \hline
     MaxVit-T & Parallel & 30,533,309 & 5,455,929,088 \\
     \hline
     MaxVit-T & Cascaded & 30,542,913 & 5,455,938,692 \\
     \hline \hline
     ConvNext & No & 87,767,364 & 15,372,622,848 \\
     \hline
     ConvNext & Parallel & 87,817,589 & 15,372,673,024 \\
     \hline
     ConvNext & Cascaded & 87,827,193 & 15,372,682,628 \\ 
     \hline  
    \end{tabular}}
    \label{tab:modealparameters}
\end{table}

\section{Investigated MTL-based approaches for VMMC}
\label{sec:methodology}

\subsection{Architectures}
\label{sec:architectures}


To fully exploit the hierarchical structure of datasets which offer multi-label stratified annotations in the form of super- and subclasses, we investigate a multi-task approach, which combines the two classification objectives: car make and car model recognition.  
In this work, we learn from homogeneous input datasets, in the form of RGB images that we resize to 224x224. The tackled tasks are also homogeneous, as we aim to perform two supervised classifications. 


\begin{table*}[ht]
    \centering
    \begin{tabular}{|c|c|c|c|c|c|c|}
        \hline
        \multirow{2}{*}{Ref} & \multicolumn{6}{c|}{Dataset} \\ \cline{2-7}
        & Name & \#Makes & \#Models & \#Images & Image Size & Ground Truth \\ \hline
        \cite{avianto2022cnn} & InaV-Dash & 4 & 10 & 4,192 & 1920 $\times$ 1080 & Semi-automatic \\ \hline
        \cite{lyu2022framework} & DVMM & 23 & 326 & 228,463 & 256 $\times$ 256 & Semi-automatic \\ \hline
        \cite{krause2015fine} & Stanford Cars & 49 & 196 & \makecell{16,185 \\ (8,144 with public labels)} & 360 $\times$ 240 & Manual \\ \hline
        \cite{yang2015large} & CompCars & 163 & 1,716 & 186,726 & Varies & Manual \\ \hline
    \end{tabular}
    \caption{Comparison of related work for MTL for make and model vehicle recognition - datasets}
    \label{tab:VMMRcomparisonData}
\end{table*}

We start from three ubiquitous architectures: DenseNet121 \cite{huang2017densely}, MaxVit T \cite{tu2022maxvit}, and ConvNext Base \cite{liu2022convnet}. Our choice was based on two main factors. First of all, we wanted to measure the influence of MTL on different model families, thus we included a CNN and a Transformer model. ConvNext is a special case, as it is a modern CNN, which managed to achieve performances which rival transformer-based models by emulating some of their characteristics (i.e., increasing field of view, the usage of the inverted bottleneck). The second criterion used for the choice of base model was the performance obtained during an initial experiment on the Stanford Cars dataset. This allowed us to select the most suitable model for this problem, from each category.  

Each model is explored in three forms. In the first one, the vanilla model is used as single-task classier able to predict the car model. The second and third ones approach the multi-task architecture. The former introduces a parallel MTL objective, in the form of a classification head tasked with predicting the car's make. 
The cascaded model first performs the car make classification task. The features used for this prediction are then concatenated to the logits obtained by the super-class classification head. This combined representation is used for the decision on the more granular task of predicting the car model.





In our work, we will use the \textit{base model} term, which refers to the entire model in the single-task configuration, and to the model without its final classification layer in the multi-task configuration. The MTL model adds a classification head: a fully-connected layer (\textbf{FCL}) for each task, to the base model, as seen in \textbf{Table} \ref{tab:MMVRcomparison}. As mentioned above, depending on the relation between these heads we can either have parallel heads or cascaded heads.

All of our models are trained end-to-end starting from ImageNet weights, with both objectives optimized simultaneously throughout the entire training process. This contrasts with previous approaches (i.e., \cite{lyu2022framework}), where heads are trained either sequentially or simultaneously for a limited number of steps followed by a secondary phase in which most parameters are frozen while only the main head continues training.


In both MTL approaches (cascade and parallel), a weighted joint loss was used. Taking into account that we deal with $m = 2$ classification tasks, we used the cross-entropy for both model loss and make loss and we transformed the equation \ref{eq:loss} into:

\begin{equation} 
    \begin{split}
    err_z(h) =  & \lambda_1 \frac{1}{n}\sum_{i=1}^{n} loss(h_1(image^i),model^i)  + \\
                & \lambda_2 \frac{1}{n} \sum_{i=1}^{n} loss(h_2(image^i),make^i) 
    \end{split}
    \label{eq:WeightedLoss}
\end{equation}

\begin{table}[ht]
    \centering
    \scriptsize
    \caption{Test accuracy for car model prediction, no dropout on Stanford Cars.}
    \scalebox{1.25}{ 
    \begin{tabular}{||c c c c||} 
     \hline
     \textbf{Model} &  \textbf{Multi-Task} & \textbf{Loss Weights}  & \textbf{Acc.} \\
     \hline\hline
     DenseNet-121 & No & - & 0.833  \\ 
     \hline
     DenseNet-121 & Parallel & [0.9, 0.1] & \textbf{0.861} \\
     \hline
     DenseNet-121 & Cascaded & [0.9, 0.1] & 0.842 \\
     \hline
     DenseNet-121 & Parallel & [0.5, 0.5] & 0.833 \\
     \hline
     DenseNet-121 & Cascaded & [0.5, 0.5] & 0.826 \\
     \hline
     DenseNet-121 & Parallel & [0.2, 0.8] & 0.787  \\
     \hline
     DenseNet-121 & Cascaded & [0.2, 0.8] & 0.776\\
     \hline\hline
     MaxVit-T & No & - & 0.878 \\ 
     \hline
     MaxVit-T & Parallel & [0.9, 0.1] & \textbf{0.882}  \\
     \hline
     MaxVit-T & Cascaded & [0.9, 0.1] & 0.881 \\
     \hline
     MaxVit-T & Parallel & [0.5, 0.5] & 0.866  \\
     \hline
     MaxVit-T & Cascaded & [0.5, 0.5] & 0.864 \\
     \hline
     MaxVit-T & Parallel & [0.2, 0.8] & 0.871 \\
     \hline
     MaxVit-T & Cascaded & [0.2, 0.8] & 0.872 \\
     \hline \hline
     ConvNext & No & - & \textbf{0.883} \\ 
     \hline
     ConvNext & Parallel & [0.9, 0.1] & 0.873  \\
     \hline
     ConvNext & Cascaded & [0.9, 0.1] & 0.871 \\
     \hline
     ConvNext & Parallel & [0.5, 0.5] & 0.875  \\
     \hline
     ConvNext & Cascaded & [0.5, 0.5] & 0.861 \\
     \hline
     ConvNext & Parallel & [0.2, 0.8] & 0.842  \\
     \hline
     ConvNext & Cascaded & [0.2, 0.8] & 0.808 \\
     \hline  
    \end{tabular}}
    \label{tab:experiment1}
\end{table}

In \textbf{Table} \ref{tab:modealparameters}, we present the number of parameters of the base models and their parallel and cascaded multi-task variants. 
In addition to these models, other established architectures were tested, including: MobileNet \cite{howard2017mobilenets} \cite{sandler2018mobilenetv2} \cite{howard2019searching}, ResNet \cite{he2016deep}, VGG \cite{simonyan2014very}, EfficientNet \cite{tan2019efficientnet}, Swin \cite{liu2021swin}, and ViT \cite{dosovitskiy2020image}. 

However, in our initial experiments with one of the datasets, they underperformed when compared to the three chosen base models by at least 5\% accuracy score. This trend continued even after the addition of the multi-task extensions.

\section{Experiments}
\label{results}

In this section, we will provide additional information regarding our experimental setup including details about: the hardware, dataset characteristics, and hyper-parameters choices. 

\subsection{Dataset}

Stanford Cars \cite{krause2015fine}, consists of 16,185 images split into two sets: 8,144 training samples and 8,041 testing samples. Each image contains a single car for which three labels are provided: the car's make, model, and year of fabrication. The samples have varying resolutions and most of the images being approximately 360 × 240 pixels. 

In the dataset, there are 196 classes for the model classification problem, which are part of 49 car-make classes. The distribution is balanced in both the training and the test sets. However, the test set labels are not publicly available. We conducted our experiments on the 8144 training samples, which we divide into three sets: training, validation, and test with the ratios 0.7, 0.2 and 0.1, as we were unable to find any public submission page which would allow us to measure the performance on the original testing samples.

\begin{table}[ht]
    \centering
    \scriptsize
    \caption{Test accuracy for car model prediction, weights ($\lambda_1=0.9$, $\lambda_2=0.1$). Drop. indicates the dropout value between the feature encoder and the fully-connected decision layers.}
    \scalebox{1.25}{ 
    \begin{tabular}{||c c c c c||} 
     \hline
     \textbf{Model} &  \makecell{\textbf{Multi}\\\textbf{Task}} & \textbf{Drop.}  & \makecell{\textbf{Stanford}\\\textbf{Acc.}}  & \makecell{\textbf{Comp}\\\textbf{Acc.}} \\
     \hline\hline
     DenseNet-121 & No & 0 & 0.833  & 0.723 \\
     \hline
     DenseNet-121 & Parallel & 0 & \textbf{0.861}  & 0.743 \\
     \hline
     DenseNet-121 & Cascaded & 0 & 0.842  & 0.734 \\
     \hline
     DenseNet-121 & No & 0.25 & 0.832  & 0.798  \\ 
     \hline
     DenseNet-121 & Parallel & 0.25 & 0.841  & 0.806 \\
     \hline
     DenseNet-121 & Cascaded & 0.25& 0.835  & 0.810 \\
     \hline
     DenseNet-121 & No & 0.5 & 0.832 & 0.724 \\ 
     \hline
     DenseNet-121 & Parallel & 0.5 & 0.844 & 0.739 \\
     \hline
     DenseNet-121 & Cascaded & 0.5 & 0.830 & 0.735 \\
     \hline\hline
     MaxVit-T& No & 0 & 0.878 & 0.766 \\ 
     \hline
     MaxVit-T & Parallel & 0 & \textbf{0.882} & 0.778 \\
     \hline
     MaxVit-T & Cascaded & 0 & 0.881  & 0.781 \\
     \hline
     MaxVit-T & No & 0.25 & 0.873  & 0.765 \\ 
     \hline
     MaxVit-T & Parallel & 0.25 & 0.864  & 0.775 \\
     \hline
     MaxVit-T & Cascaded & 0.25 & 0.877  & 0.778 \\
     \hline
     MaxVit-T & No & 0.5 & 0.88  & 0.722 \\ 
     \hline
     MaxVit-T & Parallel & 0.5 & 0.869  & 0.738 \\
     \hline
     MaxVit-T & Cascaded & 0.5 & 0.873  & 0.734 \\
     \hline \hline
     ConvNext & No & 0 & 0.883  & 0.754  \\ 
     \hline
     ConvNext& Parallel & 0 & 0.873  & 0.777 \\
     \hline
     ConvNext& Cascaded & 0 & 0.871  & 0.760 \\
     \hline
     ConvNext& No & 0.25 & \textbf{0.894}  & 0.771 \\ 
     \hline
     ConvNext& Parallel & 0.25 & 0.861 & 0.797 \\
     \hline
     ConvNext& Cascaded & 0.25& 0.862 & 0.772 \\
     \hline
     ConvNext& No & 0.5 & 0.866  & 0.749 \\ 
     \hline
     ConvNext& Parallel & 0.5 & 0.868  & 0.780 \\
     \hline
     ConvNext& Cascaded & 0.5 & 0.857  & 0.757 \\
     \hline  
    \end{tabular}}
    \label{tab:experiment2}
\end{table}

Stanford Cars provides images captured in real scenes, with different angles and lighting conditions. These variables change even in the same class, resulting in a significant intra-class variation. Another challenge is the inter-class similarity as different models from the same manufacturers can have similar shapes, especially from certain angles. Despite the difficulties, the hierarchical structure of the dataset and its "captured-in-the-wild" characteristics,  which reflect real-world, unconstrained conditions, make the dataset a desirable one for measuring the effects of multitask learning. 

CompCars \cite{yang2015large} is a large-scale image dataset designed for complete car analysis. It contains car images obtained from the web and surveillance cameras from five angles with annotations for car make, model, color, and year. Furthermore, car parts are also labeled, but in our study such information is not needed. For our problem, we can use 30,955 images using the split from the dataset: 12,813 train images, 3203 validation images, and 14,939 test images. All sets have samples from 164 super-classes (car brands) and 4446 sub-classes (car models). To be consistent with Stanford Cars, we consider the same car models from different years as different sub-classes. This increases the difficulty, as the initial number of car models is only 1716. 

As we already mentioned in Section \ref{sec:VMMR}, other datasets have been used for VMMC problem. A comparison of these datasets is presented in \textbf{Table} \ref{tab:VMMRcomparisonData}.

We have tested our MTL approaches on StanfordCar and CompCar datasets only because: (1) the ground-truth was built manually for these two datasets (except for InaV-Dash and DVMM where semi-automatic methods are used in the annotation process), (2) they contain a large number of classes for both tasks (make and model classification) with (3) a high diversity in vehicle makes and models, suitable for fine-grained classification. In addition to offering a large number of images and a wide range of vehicle makes and models (providing a comprehensive dataset for testing multi-task approaches), Stanford Cars and CompCars are both extensively used in the research community and serve as reliable benchmarks for comparing the performance of various models. CompCars may be used to test models in a variety of real-world circumstances because it incorporates photos from both web-nature and surveillance-nature scenarios. 
Furthermore, the DVMM dataset was not explicitly made for vehicle model and make identification (instead, it focuses more on image splicing and multimedia analysis), while the InaV-Dash dataset is characterized by a restricted diversity of vehicle makes and models (mainly focused on Indonesian automobiles). 

\subsection{Numerical results}

During our experiments, we resize the images to a resolution of 224x224. We train all models for 25 epochs, a value determined empirically which seems to be a point from which no further improvements are obtained with our algorithms. 

\begin{figure*}[hbt!]
\centering
\centerline{\includegraphics[scale=0.65]{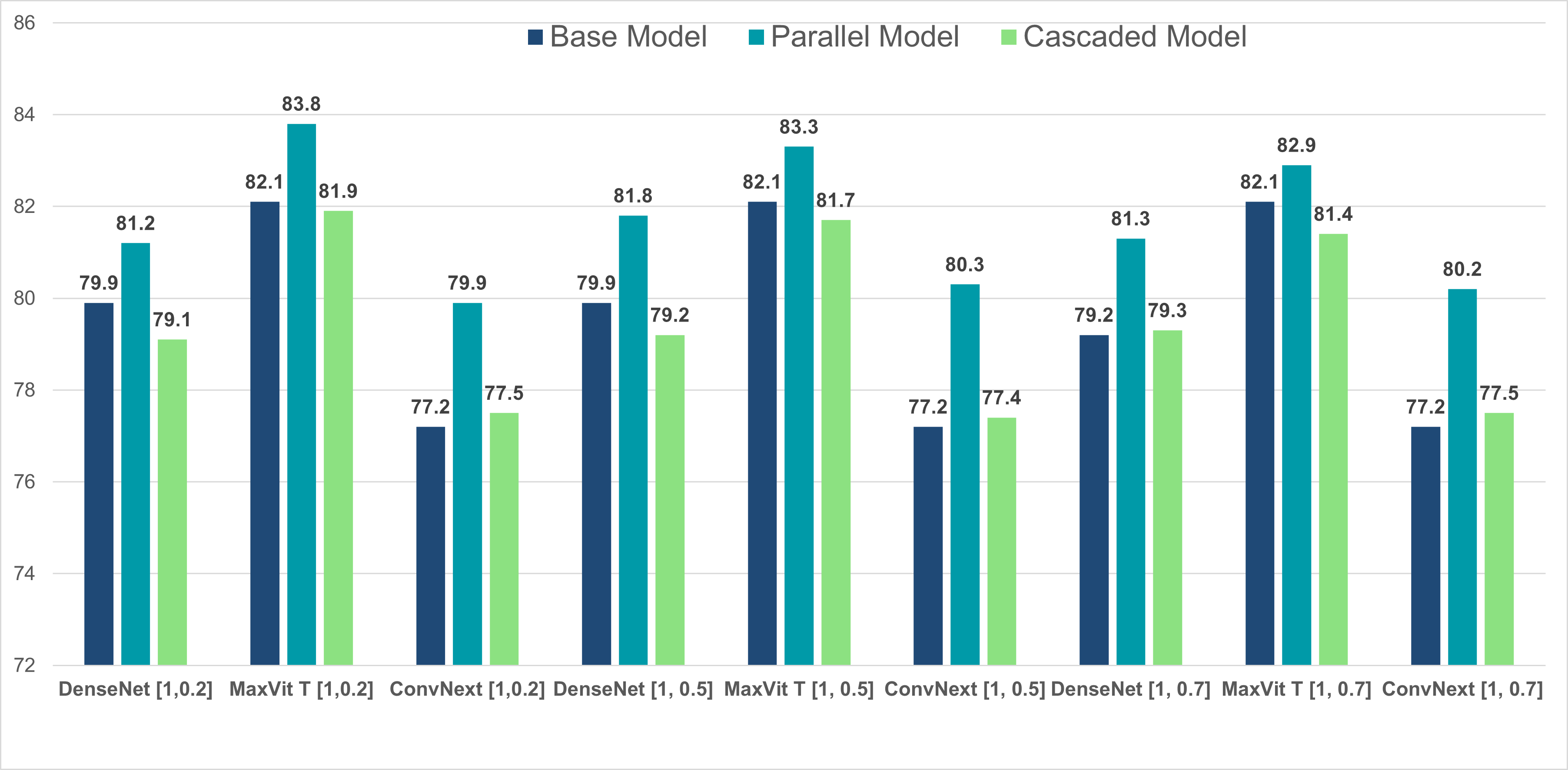}}
	\caption{Performance on CompCars test set 
    for make prediction depending on MTL model and architecture choice when dropout is set to 0.5. For each pair of lambda values and each investigated model we present three bars describing the performance of each combination of tasks: ST, parallel MTL and cascaded MTL.}
	\label{fig:performance_dropout05}
\end{figure*}

The multi-task objective is optimized using Adam, starting from a learning rate of 3e-4 that is scheduled following the one-cycle policy \cite{smith2019super}. In this algorithm, the learning rate gradually increases from an initial value to a maximum value, followed by a gradual decrease, which becomes smaller than the initial learning rate value. 

In our investigation, a number of parameters vary from one experiment to another. Two of these variables are the loss weights $\lambda_1$ and $\lambda_2$ used to balance the car model classification objective with the additional output of the car make prediction. 

Another important value which changes through our experiments is the dropout rate applied before the final prediction. It is interesting to observe the effects of the dropout, especially when used in tandem with the MTL paradigm, as both have a regulative effect on the learning process. 

Work in progress:
Evaluation metric: We compute multi-tasking performance of method m as the average per-task drop with respect to the single-tasking baseline b (i.e different networks
trained for a single task each):

\subsubsection{Experiment1 - choice of the baselines}
\label{sec:exp1}

We introduced our variables step by step to carefully gauge the effect of the MTL architecture in isolation. 
Thus, the first experiment does not use augmentations or dropout layers, and the objective weights are set to 0.9 for the main task and 0.1 for the coarser secondary classification. Thus, the first experiment does not dropout layers, and the objective weights are set to $\lambda_1$=0.9 for the main task and $\lambda_2$=0.1 for the coarser secondary classification.
The results obtained by the the forms of the investigated models on the Stanford Cars dataset are depicted in \textbf{Figure} \ref{fig:experiment0}.  We can notice that, from the list of examined models, DenseNet121 is the best suited foundational CNN for the StanfordCar dataset. When comparing the three transformer-based architectures \cite{vaswani2017attention}, MaxVit Tiny shows the highest potential. ConvNext also proves itself, again, as a powerful backbone which can rival any competing backbone.

The same procedure is conducted on the CompCars dataset, with the results presented in \textbf{Figure} \ref{fig:experiment0_compcars}. 
We observe that EfficientNetB0 and DenseNet121, both foundational CNN architectures, exhibit similar performance levels. Among transformer-based models, MaxVit Tiny consistently stands out as the top performer.

Based on these initial results, we decided to continue our experiments with MaxVit Tiny and both DenseNet and ConvNext as the latter introduces numerous changes to the architecture of traditional CNN in an effort to emulate the advantages of transformers. 

\subsubsection{Experiment2 - searching the best loss weights}
\label{sec:exp2}

We continue by varying the weights that balance the importance of the two tasks. We choose 3 pairs of weights that cover the full spectrum: significantly more importance to the main task ($\lambda_1 = 0.9$, $\lambda_2 = 0.1$), equal importance ($\lambda_1 = 0.5$, $\lambda_2 = 0.5$), and more importance to make classification ($\lambda_1 = 0.2$, $\lambda_2 = 0.8$). 
In this experiment, no dropout is used.
The results for the StanfordCars dataset are shown in \textbf{Table } \ref{tab:experiment1}.

\begin{figure*}[hbt!]
\centering
\centerline{\includegraphics[scale=0.65]{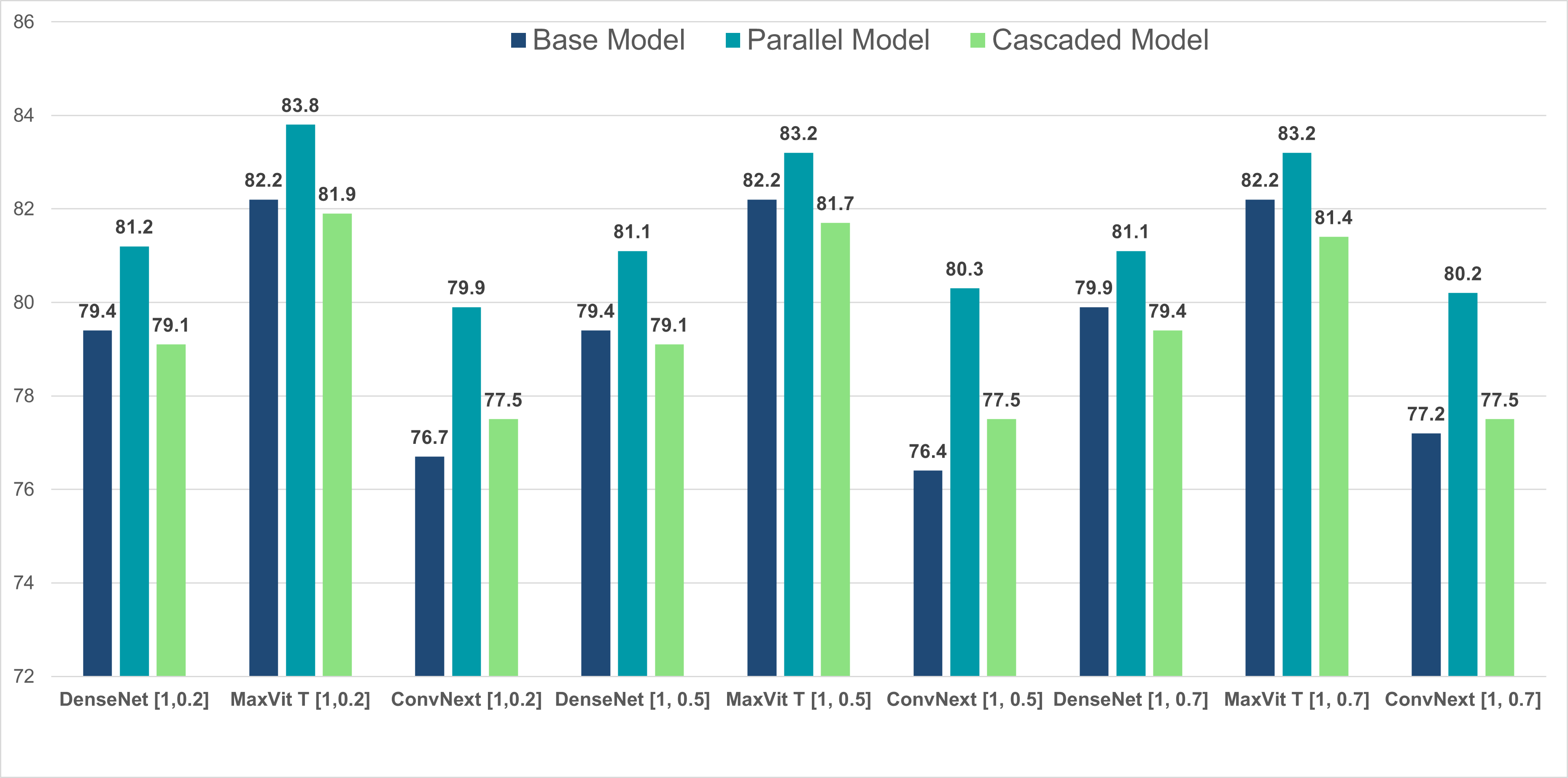}}
	\caption{Performance on CompCars test set 
    depending on MTL model and architecture choice when dropout is set to 0.25.A set of three bars is showcased for the investigation combinations of models and lambda values.The first bar describes the performance of the ST model, while the second and third show the accuracy of the parallel and cascaded MTL variants.}
	\label{fig:performance_dropout025}
\end{figure*}

Certain trends become apparent when analysing the results. First, parallel MTL usually outperforms cascade variation for all three architectures. Secondly, as one might expect, as more importance is given to the model classification, the performance increases. A notable expectation is in the case of MaxVit-T where using weights favor the more course objective, the granular performance also increases, when compared to the prospect of using equal weights.
Lastly, MTL yields better results when applied on DenseNet and MaxVit; however, based on the numerical results of the experiments, it seems to hinder the performance of ConvNext.

\subsubsection{Experiment3 - dropout}
\label{sec:exp3}

As the previous experiment suggests, the use of the weights ($\lambda_1 = 0.9$, $\lambda_2 = 0.1$) leads the MTL models to the best performance. An outlier is present in the form of the parallel ConvNext with equal weights, but since the difference is rather small and the MTL does not lead to the best performance in the model's case, we considered it acceptable to set the weights to the pair, which generally leads to better accuracy. In the next experiment, we introduced the dropout variable giving it three values: 0, 0.25 and 0.5. Although the experiments without dropout are equivalent to the ones from the previous experiment (see Table \ref{tab:experiment1}), we will also present them in \textbf{Table }\ref{tab:experiment2} to facilitate comparison of various models on the StanfordCar dataset.

Combining MTL with dropout does not increase the performance of the models in the VMMC task. Looking at the full list of results, we can observe that parallel MTL with no dropout obtains the best results for DenseNet and MaxVit-T. The presence of a 0.25 dropout increases the accuracy of ConvNext, but this result is again without the additional objective. Another aspect which is worth mentioning is the fact that in the case of MaxVit-t with dropout the cascaded architecture obtains better results than the parallel one.

\subsubsection{Experiment4 - car make classification}
\label{sec:exp4}
In our next experiments, we also focus on the accuracy results of the car-make classification problem. Moreover, we experiment again with the task weights, but this time we fix the main's objective importance to $\lambda_1 = 1.0$ and test different values for the car make weight $\lambda_2 \in \{0.2, 0.5, 0.7\}$.
The dropout rate was set to 0.25 or to 0.5.

Other trends emerge from the experiments presented in \textbf{Table} \ref{tab:experiment3}. The more coarse objective, car make classification, benefits from the prescience of the parallel MTL training procedure. As the base model only predicts the model of the car, we obtain the make by extracting the manufacturer from that prediction. As multitask architectures directly learn the make label, it is not surprising that it over-performs in this scenario. This is especially true for the DenseNet architecture. 


\newcommand{\multilinecell}[1]{%
    \begin{tabular}{@{}c@{}} #1 \end{tabular}%
}



\begin{table}[ht]
    \centering
    \scriptsize
    \caption{Test accuracy for car model and make prediction. In each accuracy cell, the first line showcases the performance of the  ST method, while the second and third lines contain the accuracy for the parallel and cascaded MTL variants. Result obtained on the Stanford Cars dataset.}
    \scalebox{1.2}{ 
    \begin{tabular}{||c c c c c||} 
     \hline
     \textbf{Model} &  \textbf{\multilinecell{Loss \\ Weights}} & \textbf{Dropout}  & \textbf{\multilinecell{Model \\ Acc.}} & \textbf{\multilinecell{Make \\ Acc.}} \\
     \hline\hline
     DenseNet-121 & [1, 0.2] & 0.25 & \multilinecell{0.858 \\ \textbf{0.878} \\ 0.858} & \multilinecell{0.923 \\0.924\\ 0.908}\\
     \hline
     DenseNet-121 & [1, 0.5] & 0.25 & \multilinecell{0.858 \\ 0.869 \\ 0.859} & \multilinecell{0.923 \\\ \textbf{0.928} \\ 0.920}\\
     \hline
     DenseNet-121 & [1, 0.7] & 0.25 & \multilinecell{0.858 \\ 0.872 \\ 0.862} & \multilinecell{0.923 \\ \textbf{0.928} \\ 0.918}\\ 
     \hline
     DenseNet-121 & [1, 0.2] & 0.5 & \multilinecell{0.860 \\ 0.864 \\ 0.877} & \multilinecell{0.915 \\0.921\\ 0.921}\\   
     \hline
     DenseNet-121 & [1, 0.5] & 0.5 & \multilinecell{0.860 \\ 0.873 \\ 0.865} & \multilinecell{0.927 \\ \textbf{0.928} \\ 0.924}\\  
     \hline
     DenseNet-121 & [1, 0.7] & 0.5& \multilinecell{0.860 \\ 0.872 \\ 0.864} & \multilinecell{0.934 \\0.923\\ 0.916}\\  
     \hline \hline
     MaxVit-T & [1, 0.2] & 0.25 &\multilinecell{\textbf{0.895} \\ 0.893 \\ 0.887} & \multilinecell{0.949 \\ 0.950 \\ 0.938}\\  
     \hline
     MaxVit-T & [1, 0.5] & 0.25 & \multilinecell{\textbf{0.895}  \\ 0.894 \\ 0.887} & \multilinecell{0.949\\ 0.950 \\ 0.949}\\  
     \hline
     MaxVit-T & [1, 0.7] & 0.25 & \multilinecell{\textbf{0.895}  \\ 0.889 \\ 0.889 } & \multilinecell{0.949 \\ 0.950 \\ 0.943}\\  
     \hline
     MaxVit-T & [1, 0.2] & 0.5 & \multilinecell{0.892 \\ 0.891\\ 0.891} & \multilinecell{0.951 \\ 0.953\\ 0.936} \\ 
     \hline
     MaxVit-T & [1, 0.5] & 0.5 & \multilinecell{0.892 \\ 0.883 \\ 0.885 } & \multilinecell{0.951 \\0.945 \\ 0.947}\\  
     \hline
     MaxVit-T& [1, 0.7] & 0.5&  \multilinecell{0.892 \\ 0.894 \\ 0.865 } & \multilinecell{0.951 \\ \textbf{0.953} \\ 0.948}\\  
     \hline \hline
     ConvNext & [1, 0.2] & 0.25 & \multilinecell{\textbf{0.895} \\ 0.894 \\ 0.865 } & \multilinecell{0.952 \\ \textbf{0.953} \\ 0.948}\\   
     \hline
     ConvNext & [1, 0.5] & 0.25 & \multilinecell{\textbf{0.895} \\ 0.881 \\ 0.889 } & \multilinecell{0.952 \\0.933 \\ 0.938}\\ 
     \hline
     ConvNext & [1, 0.7] & 0.25 & \multilinecell{\textbf{0.895}\\ 0.883 \\ 0.888 } & \multilinecell{0.952 \\0.941\\ 0.945}\\ 
     \hline
     ConvNext & [1, 0.2] & 0.5 &  \multilinecell{0.886 \\0.861 \\ 0.861} &  \multilinecell{0.948 \\0.921 \\ 0.913} \\ 
     \hline
     ConvNext & [1, 0.5] & 0.5&  \multilinecell{0.886 \\ 0.888 \\ 0.879 } & \multilinecell{0.948 \\0.935 \\ 0.949}\\
     \hline
     ConvNext & [1, 0.7] & 0.5 & \multilinecell{0.886 \\ 0.889 \\ 0.877 } & \multilinecell{0.948 \\0.942 \\ 0.942}\\ 
     \hline \hline  
    \end{tabular}}
    \label{tab:experiment3}
\end{table}

\begin{figure*}[hbt!]
\centering
\centerline{\includegraphics[scale=0.65]{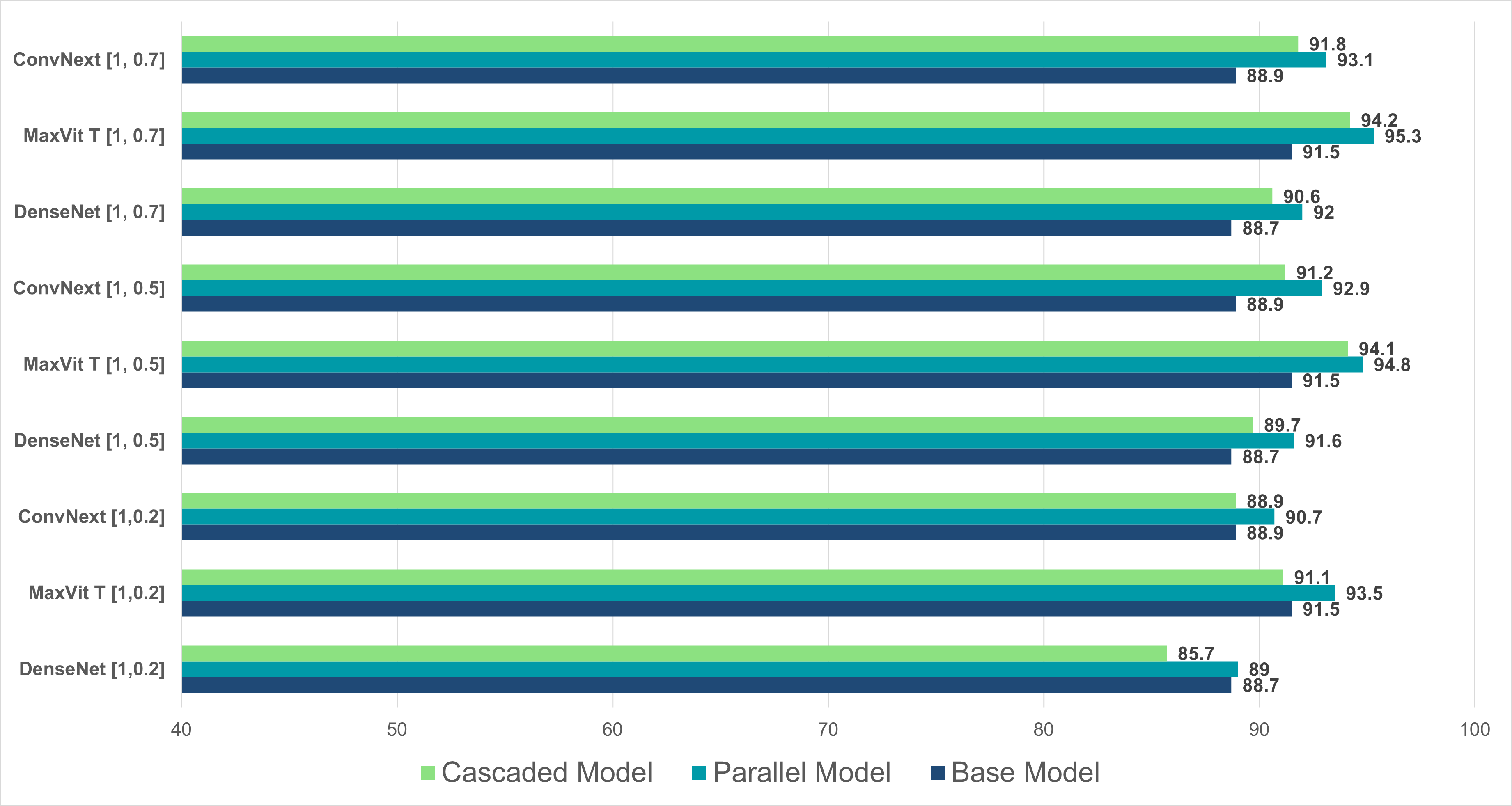}}
	\caption{Performance on CompCars test set for make prediction
    depending on MTL model and architecture choice when dropout is set to 0.5.}
	\label{fig:performance_dropout05_make}
\end{figure*}

\begin{figure*}[hbt!]
\centering
\centerline{\includegraphics[scale=0.65]{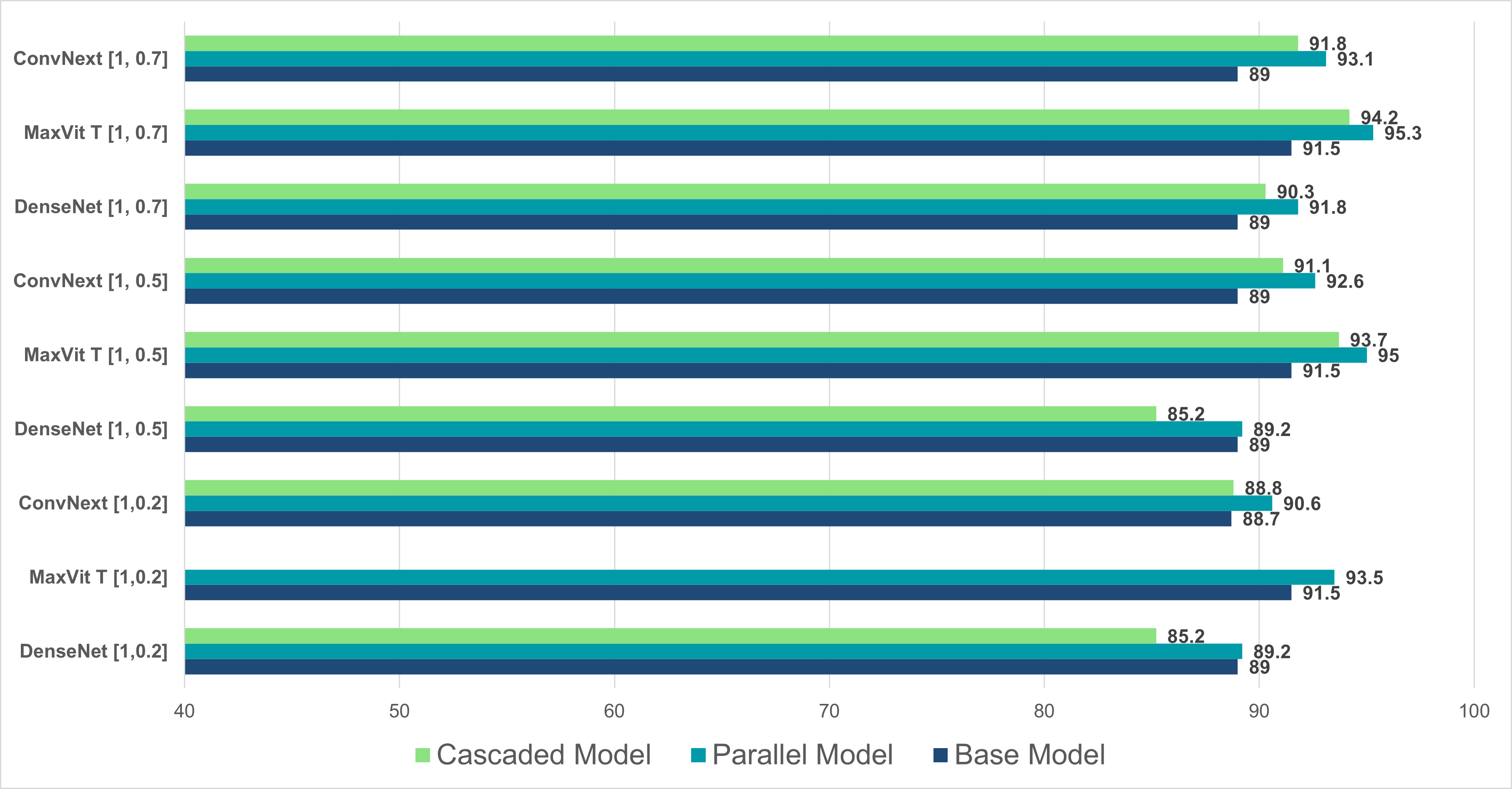}}
	\caption{Performance on CompCars test set for make prediction
    depending on MTL model and architecture choice when dropout is set to 0.25.}
	\label{fig:performance_dropout025_make}
\end{figure*}

The model classification performance of equivalent experiments performed on the CompCars dataset are presented in \textbf{Figure} \ref{fig:performance_dropout05} and \textbf{Figure} \ref{fig:performance_dropout025}. The former showcases results when the dropout is set to 0.5, while the latter illustrates the performance when a value of 0.25 is used for the same hyperparameter. We also present the accuracy for the make classification task in \textbf{Figure} \ref{fig:performance_dropout05_make} and \textbf{Figure} \ref{fig:performance_dropout025_make}. All figures follow the same convention, in which three bars are used to show the performances of the single-task (base model), parallel multi-task and cascaded multi-task for each combination of model architecture (DenseNet, MaxVit, ConvNext) and choice of weights.

All of our experiments were conducted in the Google Colaboratory environment using the NVIDIA A100 Tensor Core GPU.  

\subsection{Discussion}



On Stanford Cars, both the base MaxVit-T and ConvNext achieve a top accuracy of 0.895. This is an improvement over the best performances obtained when no augmentations were present, especially for the transformer model. An MTL model derived from each MaxVit and ConvNext competes with the best performing base forms, achieving 0.894 accuracy, namely, the parallel ConvNext with dropout 0.25, weighted [1, 0.2] and the parallel MaxVit with the same dropout with task weights [1, 0.5]. As the advantage of the MTL framework is not evident in all scenarios when the Stanford Cars dataset is used, we take a deeper look into the prediction by also analyzing the top-3 and top-5 accuracies.

\begin{figure*}[hbt!]
\centering
\centerline{\includegraphics[scale=0.65]{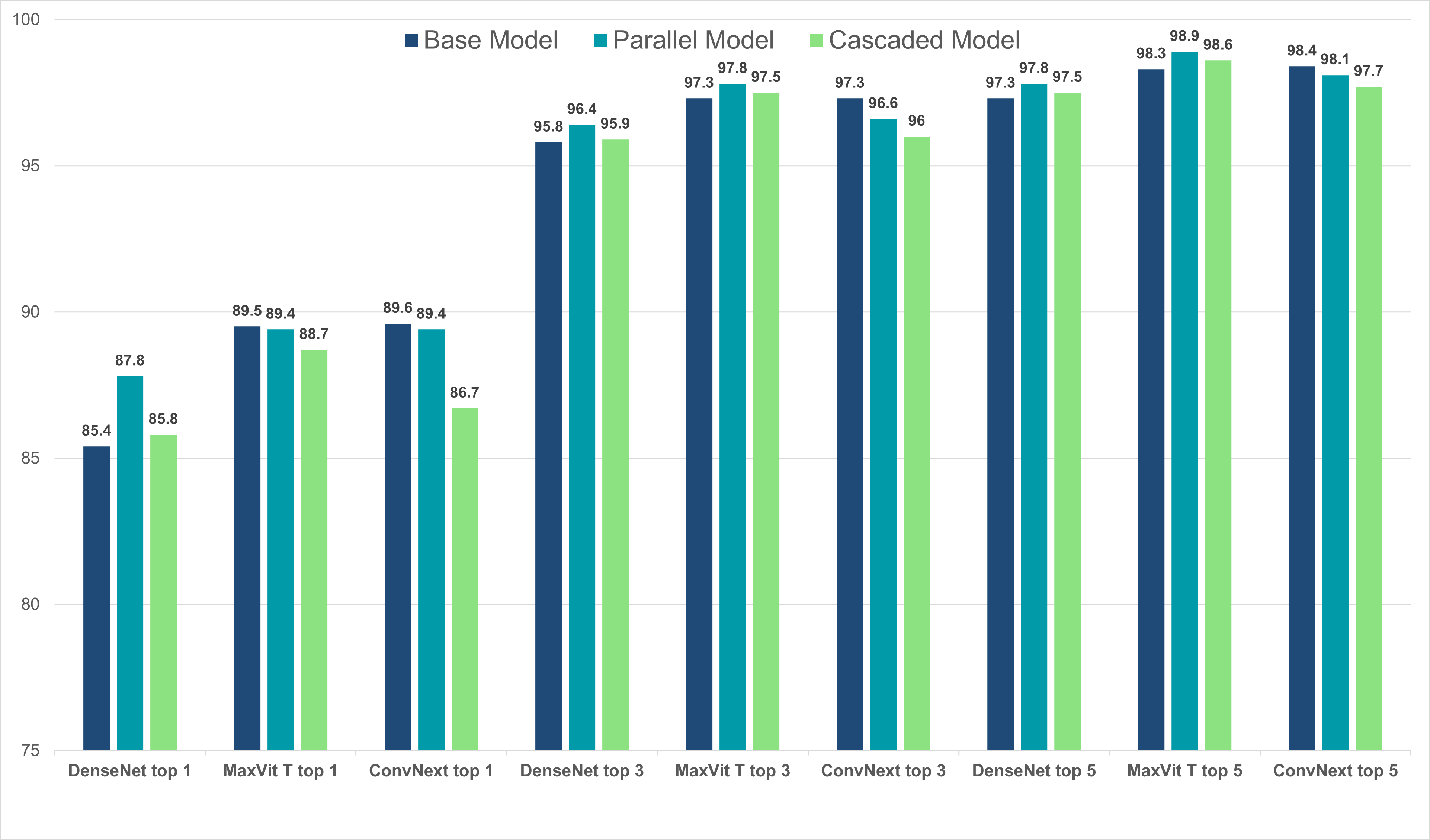}}
	\caption{Top 1, top 3 and top 5 model accuracies for DenseNet, MaxVit T and ConvNextBase in their base, parallel and cascaded form.   }
	\label{fig:topnaccuracies}
\end{figure*}

In \textbf{Fig.} \ref{fig:topnaccuracies} we showcase the performance of the best DenseNet, MaxVit and ConvNext obtained from all of our experiments on Stanford Cars (DenseNet with weights [1, 0.5] and dropout 0.5; MaxVit-T with weights [1, 0.7] and dropout 0.5 and ConvNext with weights [1, 0.2] and dropout 0.25). We go further than inspecting the pure performance and also analyse top-3 and top-5 accuracies. We run these models again on the test set and save the top-k classes with the highest confidence. For top-k accuracy, if the correct answer is found among these classes, we consider the prediction as correct.

In ConvNext case, the same behavior is present through all metrics, namely the ST model outperforms the two MT variants. Similarly, for DenseNet the ranking of the top-1 accuracy remains unchanged for top-3 and top-5. In this model's case the additional multi-task objective is beneficial, as both parallel and cascaded variants outperform the single-task variant.

The most interesting development comes from the MaxVit T models. Even though the ST model exceeds the performance of the cascaded and parallel ones in normal accuracy (top-1), for the other two metrics MT variants yield better results. Furthermore, the best top-5 accuracy is obtained by the parallel MaxVit T with a score of 98.9\% accuracy, better than DenseNet and ConvNext. This suggests that MTL embedded a broader semantic understanding, with the potential to overperform when it comes to the recognition task in ambiguous cases, while maintaining an extremely competitive top-1 accuracy.


All of our results indicate that for VMMC, the addition of a second objective usually has a beneficial effect on the performance of the model, with a small cost, in terms of the number of parameters and FLOPs. Our CNN choice, DenseNet, is the model that has gained the most from this enhancement. For the investigated vision transformer, MaxVit T, the ST and MT variants obtain almost the same top-1 accuracy score. In this scenario, the additional objectives aid in the secondary task (model recognition) and in top-5 accuracy, both suggesting a deep semantic understanding which could be less prone to overfitting. Lastly, ConvNext Base does not directly benefit from the proposed multi-task approach. We believe that this is not an illustration of the potential of symbiotic nature of this architecture and combined learning objective, but rather a manifestation of an incompatibility between a huge architecture and a small dataset.

In CompCars, reducing the dropout rate from 0.5 to 0.25 generally resulted in a slight decrease in performance for the DenseNet and ConvNext base models, while MaxVit T remained consistent or improved slightly. This suggests that MaxVit T is more robust to changes in regularization, likely due to its transformer-based design, which benefits from more active neurons during training.

Across all configurations, the Parallel Model consistently outperformed the Base Model. For example, DenseNet [1,0.2] with dropout 0.5 increased from 79.9\% to 81.2\%, and MaxVit T [1,0.2] increased from 82. 1\% to 83. 8\%. This indicates that the parallel training approach is effective in improving feature diversity and generalization.

The Cascaded Model also tended to improve over the Base Model, though typically with smaller gains compared to the Parallel Model. In several cases, such as ConvNext [1,0.7], the improvements were marginal, suggesting that cascading can add useful depth but can introduce some redundancy or instability depending on the underlying architecture.

Among the architectures, MaxVit T achieved the highest overall performance across all training modes and dropout rates, confirming its strong adaptability. DenseNet also showed reliable improvements, while ConvNext demonstrated the least benefit from either the Parallel or Cascaded strategies, potentially due to its streamlined convolutional design being less receptive to additional complexity.

Overall, the best-performing configuration was MaxVit T trained with the Parallel Model and a dropout rate of 0.25, reaching up to 83.8\% accuracy. This combination provided a strong balance between regularization and enhanced feature integration, resulting in the most robust performance.

Through our experiments and analysis, we believe that that we answered our research questions:
\begin{enumerate}
    \item[\textbf{RQ1}:] 
    The addition of cascaded and parallel multi-task learning objectives positively influences the VMMC problem by enhancing model performance through shared representations. Parallel MTL consistently outperforms the base model across all configurations. Cascaded MTL also improves performance, though typically to a lesser degree than parallel MTL, suggesting that task sequencing can be beneficial but may be more sensitive to task compatibility.

    \item[\textbf{RQ2}:] 
    The architecture of the base model plays a significant role in determining the effectiveness of the multi-task approach. Transformer-based models such as MaxVit T are able to achieve the best results using MTL, but the greatest percentage increase is found in the DenseNet architecture. 
\end{enumerate}

\section{Conclusions and Future Considerations}
\label{conclusion}

In this work, we were able to showcase the benefits of the multi-task learning paradigm for a hierarchical multi-label problem, car model, and make classification. The advantages of this technique were illustrated for both the CNN and Transformer architectures. Furthermore, in the case of the former, the MTL architecture allows the relatively small DenseNet to rival larger and more modern architectures. 

Although we were able to showcase some of the advantages of the MTL approach, we strive to provide additional analysis in the future, in order to obtain a better understanding of this framework and its best uses. Thus, we identify the following future research direction: 
\begin{itemize}
    \item Extend the analysis by adding experiments with one or more types of cross-talk multi-task architectures
    \item Analyse the performance of parallel and cascaded models when the number of task-specific layers or blocks increases
    \item Conduct test on more datasets from this field or other hierarchical ones in order to extract a more general understanding of the behaviour of MTL models
    \item Experiment and analyse the behaviour of parallel, cascaded and cross-talk multi-task and tasks different than classification, including, but not limited to semantic segmentation and detection.
\end{itemize}

\section*{Acknowledgement}
This research is supported by the project “Romanian Hub for Artificial Intelligence - HRIA”, Smart Growth, Digitization and Financial Instruments Program, 2021-2027, MySMIS no. 334906.

\balance


\printbibliography

\vfill

\end{document}